\documentclass[10pt,twocolumn,letterpaper]{article}

\usepackage{cvpr}              % To produce the CAMERA-READY version
\usepackage{graphicx}
\usepackage{amsmath}
\usepackage{amssymb}
\usepackage{booktabs}
\usepackage{amsfonts}       % blackboard math symbols
\usepackage{nicefrac}       % compact symbols for 1/2, etc.
\usepackage{microtype}      % microtypography
\usepackage{xcolor}         % colors
\usepackage{graphicx}
\usepackage{diagbox}
\usepackage{multicol}
\usepackage{enumerate}
\usepackage{times}
\usepackage{epsfig}
\usepackage{graphicx}
\usepackage{amsmath}
\usepackage{amssymb}
\usepackage{threeparttable}
\usepackage{enumitem}
\usepackage{multirow}
\usepackage{color}
\usepackage{array}

\setenumerate[1]{leftmargin = 10pt, itemsep=0pt,partopsep=0pt,parsep=-0pt,topsep=-0pt}
\usepackage[pagebackref,breaklinks,colorlinks]{hyperref}
\newcommand{\PreserveBackslash}[1]{\let\temp=\\#1\let\\=\temp}

\usepackage[capitalize]{cleveref}
\crefname{section}{Sec.}{Secs.}
\Crefname{section}{Section}{Sections}
\Crefname{table}{Table}{Tables}
\crefname{table}{Tab.}{Tabs.}

\def\cvprPaperID{5008} % *** Enter the CVPR Paper ID here
\def\confName{CVPR}
\def\confYear{2022}

\begin{document}
\makeatletter
	
	\def\hlinew#1{%
		\noalign{\ifnum0=`}\fi\hrule \@height #1 \futurelet
		\reserved@a\@xhline}
	\makeatother%
	\newcommand{\tabincell}[2]{\begin{tabular}{@{}#1@{}}#2\end{tabular}}
\title{Unsupervised Deraining: Where Contrastive Learning Meets Self-similarity}

\author{Yuntong Ye\textsuperscript{1}, Changfeng Yu\textsuperscript{1}, Yi Chang\textsuperscript{1}\footnotemark[1] , Lin Zhu\textsuperscript{2}, Xi-le Zhao\textsuperscript{3}, Luxin Yan\textsuperscript{1}, Yonghong Tian\textsuperscript{2}\\
\textsuperscript{1}School of Artificial Intelligence and Automation, Huazhong University of Science and Technology, China\\
\textsuperscript{2}School of Artificial Intelligence, Peking University, China\\
\textsuperscript{3}School of Mathematical Sciences, University of Electronic Science and Technology of China, China\\
}

\twocolumn[{%
	\renewcommand\twocolumn[1][]{#1}%
	\maketitle
	\begin{center}
  \setlength{\belowcaptionskip}{2pt}
		\centering
		\captionsetup{type=figure}
		\includegraphics[width=1.0\textwidth]{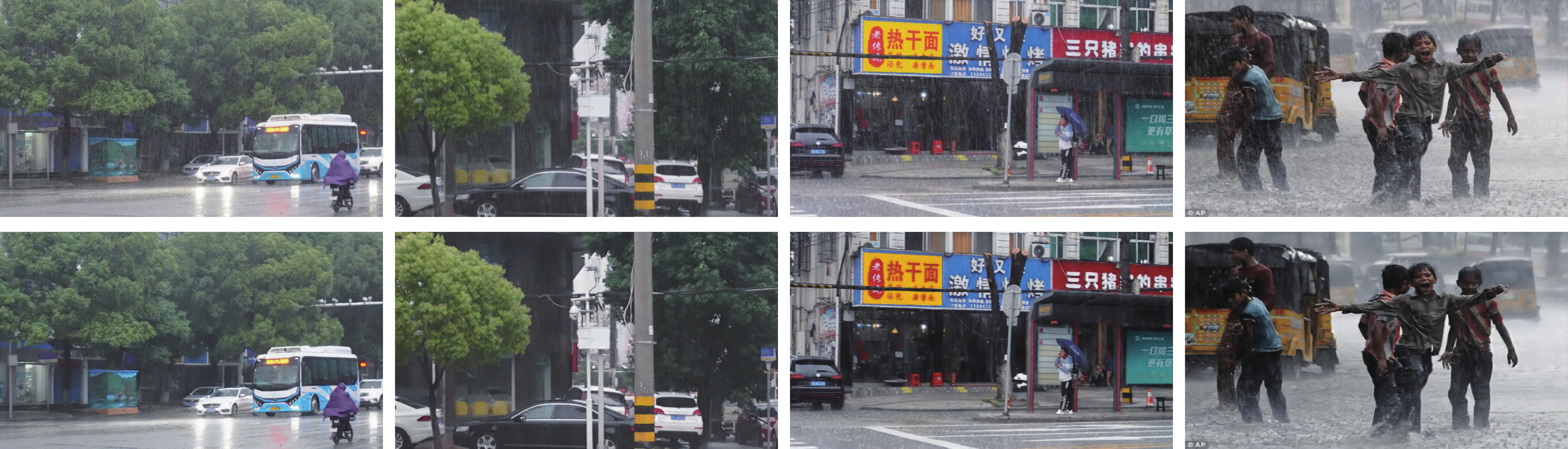}
		\captionof{figure}{The proposed method can remove both the real-world rain streaks and veiling effect meanwhile well preserve image structures in an unsupervised manner. More real results and comparisons with the state-of-the-arts can be found in the supplementary.}
	\label{fig1}
	\end{center}%
}]

\maketitle

%%%%%%%%% ABSTRACT
\begin{abstract}

~\\[-25pt]
  Image deraining is a typical low-level image restoration task, which aims at decomposing the rainy image into two distinguishable layers: clean image layer and rain layer. Most of the existing learning-based deraining methods are supervisedly trained on synthetic rainy-clean pairs. The domain gap between the synthetic and real rains makes them less generalized to different real rainy scenes. Moreover, the existing methods mainly utilize the property of the two layers independently, while few of them have considered the mutually exclusive relationship between the two layers. In this work, we propose a novel non-local contrastive learning (NLCL) method for unsupervised image deraining. Consequently, we not only utilize the intrinsic self-similarity property within samples, but also the mutually exclusive property between the two layers, so as to better differ the rain layer from the clean image. Specifically, the non-local self-similarity image layer patches as the positives are pulled together and similar rain layer patches as the negatives are pushed away. Thus the similar positive/negative samples that are close in the original space benefit us to enrich more discriminative representation. Apart from the self-similarity sampling strategy, we analyze how to choose an appropriate feature encoder in NLCL. Extensive experiments on different real rainy datasets demonstrate that the proposed method obtains state-of-the-art performance in real deraining.
\end{abstract}
\footnotetext[1]{Corresponding author}

~\\[-40pt]
%%%%%%%%% BODY TEXT
\section{Introduction}
\label{sec:intro}
The existing high-level computer vision tasks such as image segmentation \cite{chen2018encoder}, and object detection \cite{liu2016ssd} have achieved significant progress in recent years. Unfortunately, their performance would suffer from degradation under the rainy weather \cite{bahnsen2018rain, jiang2020multi, li2019rainflow}. To alleviate the influence of the rain, numerous full-supervised deraining methods have been proposed \cite{fu2017clearing, yang2017deep, zhang2018density}. Although they can achieve good results on simulated rainy image, they cannot well generalize to the real rain because of the domain gap between the simplified synthetic rain and complex real rain \cite{ye2021closing}. The goal of this work is to remove the real rain in an unsupervised manner.

To handle the real-world complex rainy images, the optimization-based methods are firstly proposed with hand-crafted priors such as the sparse coding \cite{luo2015removing}, low-rank \cite{chang2017transformed} and Gaussian mixture model \cite{li2016rain}. However, these hand-crafted priors are of limited representation ability, especially for highly complex and varied rainy scenes. To rectify this weakness, the learning-based CNN methods \cite{fu2017clearing, yang2017deep, li2018recurrent, li2019heavy} have made great progresses. The key idea of these supervised learning methods tries the best to simulate the rain as real as possible with sophisticated models, such as the additive model \cite{kang2011automatic}, screen blend model \cite{luo2015removing}, heavy rain model \cite{yang2017deep}, and comprehensive rain model \cite{hu2019depth}, to name a few. Unfortunately, there still exist gap between these synthetic rain models and real rain degradation, since the real rainy atmosphere is usually a high-order nonlinear system.

Furthermore, the semi-supervised deraining methods have been proposed to effectively improve the robustness for real rains \cite{wei2019semi, yasarla2020syn2real, wei2019deraincyclegan, ye2021closing, huang2021memory, liu2021unpaired}, where they employ the simulated labels for good initialization and unlabeled real rains for generalization. Their performances still depend on the distribution gap between the simulated and real rainy images to some extent. Once the distributions are of large distance, these semi-supervised deraining results would be less satisfactory. Very recently, the unsupervised methods have raised more attentions for real rain removal, mainly including the CycleGAN-based unpaired image translation methods \cite{zhu2019singe, jin2019unsupervised, wei2021deraincyclegan} and the optimization-model driven deep prior network \cite{yu2021unsupervised}. However, the previous methods including the unsupervised ones mainly pay attention to the property of the image or rain layer independently, yet seldom consider the mutually exclusive relationship between the two layers.

\begin{figure}[t]
  \centering
     \includegraphics[width=1.0\linewidth]{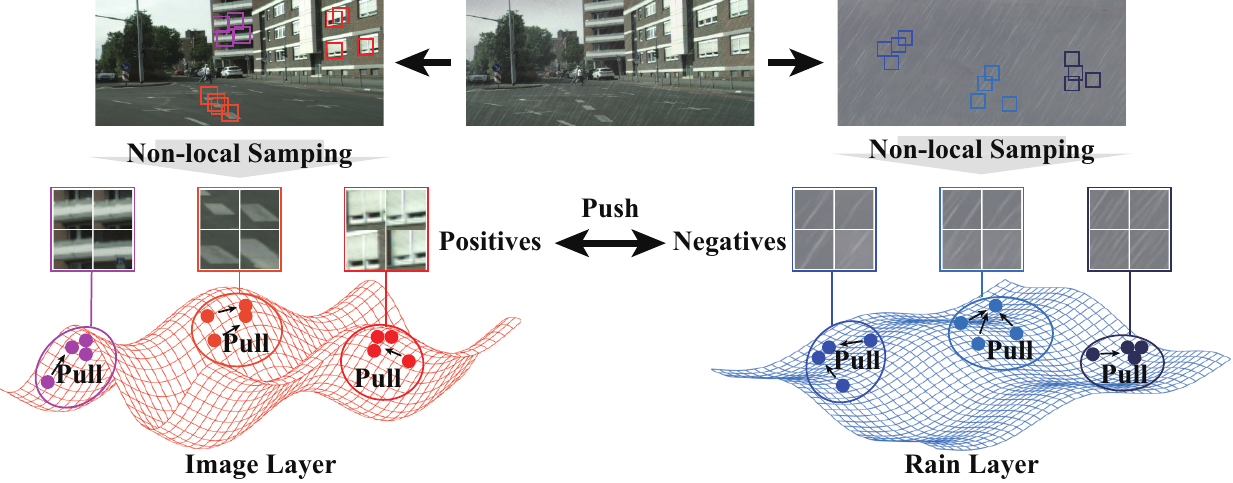}
  \caption{Most previous methods model the property of the image layer and rain layer independently in a supervised manner. In this work, we go further by considering the mutually exclusive relationship between the two layers, and propose an unsupervised non-local contrastive learning method to learn mutually exclusive relationship by pushing away the image positives and rain negatives. Moreover, the non-local self-similarity has been exploited to improve the positive/negative sampling with discriminative representation.}
   \vspace{-10pt}
  \label{Motivation}
\end{figure}

To overcome these problems, we formulate the image deraining into a contrastive learning framework \cite{chen2020simple, he2020momentum} from an unsupervised perspective. The core idea of contrastive learning is that the representation of similar samples should be pulled close together, while that of dissimilar samples should be pushed far away in the embedding space \cite{hadsell2006dimensionality, wu2018unsupervised}. Figure \ref{Motivation} illustrates the main idea of proposed method. The image deraining is formulated as an image decomposition task, in which the clean image patches are regarded as the positives while the rain layer patches as the negatives. Thus, we not only take advantage of the properties of both image and rain  layers, but also model the mutually exclusive relationship between the two layers for better decomposition. On the other hand, the proposed method does not require the clean supervision, which makes it generalize well for the real-world rainy images.

The key factor of contrastive learning is how to construct different views for both the positive and negative samples. The main stream is to augment a single instance with different transformations as the positive samples so as to learn the invariant representations \cite{chen2020simple}. However, these instance-level hand-crafted augmentations are not adequate to cover various situations. In this work, we provide a new perspective via the patch-level self-similarity within a single image. While non-local self-similarity \cite{buades2005non} has been extensively studied in the literature, this intrinsic property for capturing the cross-patch relation in a single image with contrastive learning has barely been explored for visual representation learning.

To the best of our knowledge, we are the first to incorporate non-local self-similarity into contrastive learning for positive/negative sampling. The advantage of the proposed non-local sampling is twofold. First, the non-local self-similarity sampling strategy would naturally guarantee more compact clusters for positives and negatives respectively, which would benefit us to differ the positives from negatives. Second, these positive non-local patches are  the samples searched from real images with diverse variable information, not manually generated fake samples, which would provide more faithful information for representation. Note that, the non-local strategy is not only applicable for the positive samples, but also beneficial to the negative samples. Overall, our contributions can be summarized as follow:
\begin{itemize}[leftmargin=10pt]
 % \vspace{-0.3cm}
\item We propose a non-local contrastive learning method (NLCL) for unsupervised image deraining. Compared with previous deraining methods, we not only exploit the specific property of the image and rain layers, but also model the contrastive relationship between them for better decoupling the rain layer from the clean image.
\setlength{\itemsep}{-2pt}
\item We connect the contrastive learning with the non-local self-similarity. The non-local patch sampling strategy naturally endows the positive/negative samples with more compact and discriminative representation for better decomposition. In addition, we provide an guidance of how to design a good encoder for better embedding in NLCL.
\setlength{\itemsep}{-2pt}
 \item We conduct extensive experiments on both synthetic and real-world datasets, and show that NLCL outperforms favorably state-of-the-art methods on real image deraining.
\end{itemize}

\section{Related Work}

\noindent
\textbf{Single Image Deraining.} Here, we mainly focus on the learning-based deraining methods. Most of the existing methods are full-supervised which require a large number of paired rainy and clean images as training samples \cite{yang2017deep, fu2017removing, zhu2017joint, li2018recurrent, hu2019depth, li2019heavy, wang2021multi, chen2021robust, yi2021structure, quan2021removing, fu2021rain}. Fu \emph{et al}. \cite{fu2017removing} first introduced the end-to-end residual CNN for rain streaks removal. Latter, the multi-stage \cite{yang2017deep}, multi-scale \cite{jiang2020multi}, density \cite{zhang2018density}, and attention \cite{li2018recurrent} have been widely utilized for better representation. Unfortunately, the domain gap between the complex real rain and the simplified synthetic rain would limit their generalization in real scenes. The semi-supervised deraining models \cite{wei2019semi, yasarla2020syn2real, wei2019deraincyclegan, ye2021closing, huang2021memory} could alleviate this issue to some extent. For example, Wei \emph{et al}. \cite{wei2019semi} first proposed a semi-supervised transfer learning framework via network structure sharing for real image deraining. Recently, the unsupervised deraining methods have emerged \cite{zhu2019singe, jin2019unsupervised, wei2021deraincyclegan, yu2021unsupervised}. Yu \emph{et al}.  \cite{yu2021unsupervised} connected the model-driven and data-driven methods via an unsupervised learning framework. In this work, we propose a novel contrastive learning framework for unsupervised deraining. Compared with previous methods, the NLCL could further take mutual exclusive relationship between image and rain layers into consideration.

\noindent
\textbf{Contrastive Learning.} Contrastive learning (CL) has achieved promising results in unsupervised representation learning \cite{chen2020simple, he2020momentum, chen2020improved, oord2018representation, wu2018unsupervised, chen2020big}. The main idea is to push the features of unrelated data (as negatives) and pull the related data (as positives), so as to learn the representations which are discriminative to the negatives and invariant between the positives. CL can be effectively applied by appropriately defining the positives and negatives in terms of the tasks, including the multi-views \cite{tian2019contrastive, 2020What}, temporal coherence in video sequence \cite{Han2020VideoRepresentation}, augmented transformation \cite{chen2020simple, he2020momentum}, to name a few. Recently, researches have applied the CL to low-level applications \cite{park2020contrastive, liu2021divco, wu2021contrastive}. Wu \emph{et al}. \cite{wu2021contrastive} pulled the restored image closer to ground truth (GT) and pushed them far away from the hazy image in the representation space.

Our NLCL is significantly different from \cite{wu2021contrastive} in three aspects. First, our method is completely unsupervised which does not need the GT. Second, we take the estimated image and rain layers as the positive and negative, respectively. Such an explicit disentanglement between the two layers would better facilitate us to decouple the rain from the clean image. Third, \cite{wu2021contrastive} employs a classical instance image-level samples for contrast, while we have explored the intrinsic similarity between the patches within a single image. The self-similarity within the positive or negative would further boost more compact and structural feature space.

\noindent
\textbf{Non-local Self-similarity.} The self-similarity serves as a powerful image prior model, which has been verified in various image restoration techniques, including filtering methods \cite{buades2005non, dabov2007image}, sparse optimization models \cite{mairal2009non, gu2014weighted}, and deep neural networks \cite{liu2018non, wang2018non, bell2019blind}. The nonlocal prior reveals a general image property that the similar small patches tend to recurrently appeared within a single image. This generic property could provide group sparsity of the image with structural representation. Beneficial from capturing the correlation among the self-similarity patches, these non-local based methods have achieved the state-of-the-art performances at the time, such as the BM3D in denoising \cite{dabov2007image}, WNNM in restoration \cite{gu2014weighted}, and kernelGAN in blind super-resolution \cite{bell2019blind}. In this work, we demonstrate how this intrinsic property benefits the contrastive learning in terms of the positive/negative sampling, and boosts the performance in low-level image deraining task.

\section{Non-local Contrastive Learning}
\subsection{Overview of the Framework}
\begin{figure*}[t]
  \centering
  \includegraphics[width=1.0\linewidth]{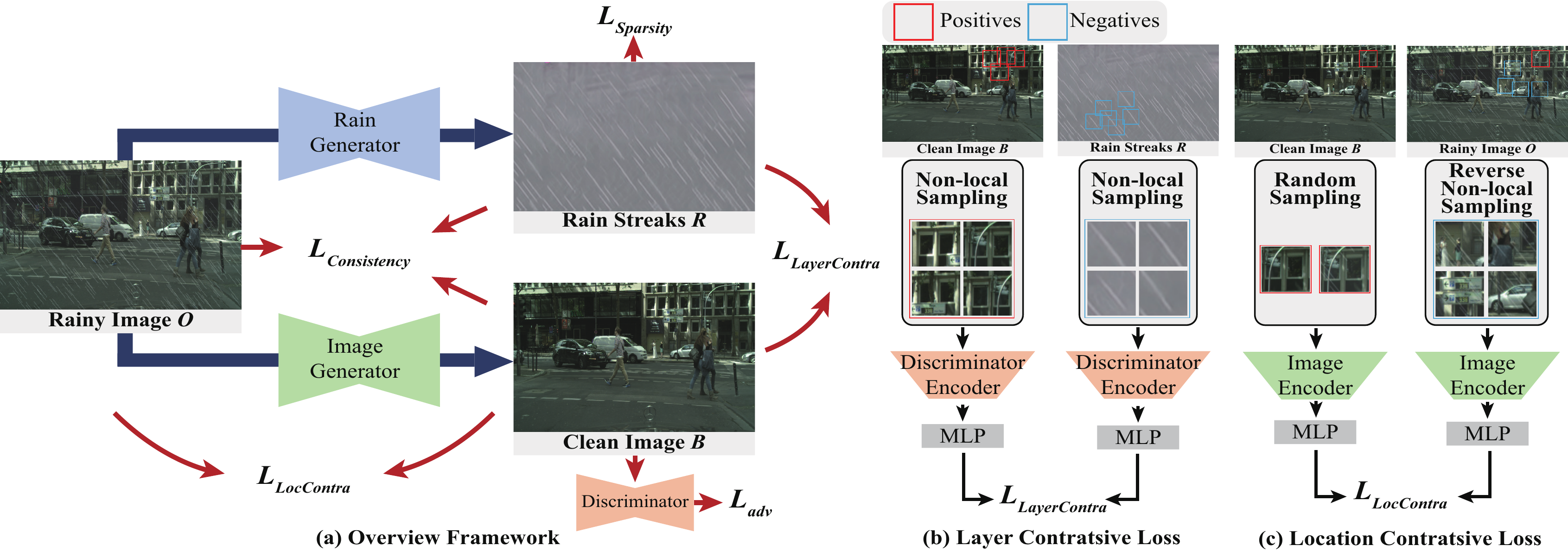}
  \caption{Overview of architecture of the proposed method. (a) The NLCL consists of two sub-networks to extract the background and the rain layers respectively with two additional contrastive constraints. (b) The layer contrastive between the clean image and rain streaks with the non-local sampling for both positives and negatives. (c) The location contrastive between the clean image and rainy image with the reverse non-local sampling for negatives.}
  \vspace{-10pt}
  \label{Structure}
\end{figure*}

Given a rainy image $\textbf{\emph{O}}$, our goal is to decompose the rainy image into a clean background layer ${\textbf{\emph{B}}}$ and a rain layer ${\textbf{\emph{R}}}$. The degradation procedure can be formulated as:
\begin{equation}\label{eq:DegradationModel}
\setlength{\abovedisplayskip}{2pt}
\setlength{\belowdisplayskip}{2pt}
\textbf{\emph{O}} = \textbf{\emph{B}} + \textbf{\emph{R}}.
\end{equation}
Note that, although we follow this simple decomposition framework \cite{li2016rain}, this does not mean the proposed method only handles the rain streak. The proposed method can well restore the heavy rain with haze or veil artifacts. Thus, the image deraining task can be formulated as an ill-posed inverse problem with following optimization function:
\begin{equation}
\setlength{\abovedisplayskip}{2pt}
\setlength{\belowdisplayskip}{2pt}
\mathcal{L}_{decom}  =  ||\textbf{\emph{B}} + \textbf{\emph{R}} - \textbf{\emph{O}}||_F^2 + \delta P_{b}(\textbf{\emph{B}}) + \lambda P_{r}(\textbf{\emph{R}}),
\label{eq:RegularizationFormulation}
\end{equation}
where the first term is self-consistent loss, namely the data fidelity term, $P_{b}$ and $P_{r}$  denote the prior knowledge for the clean image and rain streaks, respectively. Thanks to the sparsity of the rain streaks in space, in this work, we regularize the rain layer with the $L_1$ constraint: $P_{r}(\textbf{\emph{R}}) = ||\textbf{\emph{R}}||_1$ favoring the rain streaks with large discontinuities. On the other hand, for the clean images, we employ the adversarial loss \cite{Goodfellow2014Generative} to learn the distribution mapping differing the rainy image from clean image:
\begin{equation}
\begin{aligned}
\setlength{\abovedisplayskip}{2pt}
\setlength{\belowdisplayskip}{2pt}
P_{b}(\textbf{\emph{B}}) = \mathbb{E}_{\textbf{\emph{B}}}\left [ \textrm{log}D(\textbf{\emph{B}}) \right ] + \mathbb{E}_{\textbf{\emph{O}}}\left [ \textrm{log}(1-D(G_\textbf{\emph{B}}(\textbf{\emph{O}})))\right],
\end{aligned}
\end{equation}
where \emph{D} is the discriminator, and $G_\textbf{\emph{B}}$ is the generator for the clean image. The corresponding network of the decomposition-based architecture is shown in Fig. \ref{Structure}(a), which consists of two branches to restore the background ($G_{\textbf{\emph{B}}}$) and extract the rain ($G_{\textbf{\emph{R}}}$).

Most of the existing restoration methods follow the decomposition framework in Eq. (\ref{eq:RegularizationFormulation}) with different hand-crafted or learned priors, where they only consider the clean image or rain layer separately. That is to say, the Eq. (\ref{eq:RegularizationFormulation}) mainly focuses on modelling of the statistical property of the signal itself. However, it has neglected the relationship between clean image \textbf{\emph{B}}, rain layers \textbf{\emph{R}}, and observed image \textbf{\emph{O}}. In this work, we argue the relationship among these components can further help to distinguish them from each other. We introduce the contrastive learning to model the relationship between different components. Specifically, we evolve the relation $\mathcal{L}_{LayerCon}$ between the clean image \textbf{\emph{B}} and rain layer \textbf{\emph{R}}, also relation $\mathcal{L}_{LocCon}$ between clean image \textbf{\emph{B}} and observed image \textbf{\emph{O}}. Thus, the full objective function including the decomposition constraint and contrastive loss is formulated as:
\begin{equation}
\setlength{\abovedisplayskip}{2pt}
\setlength{\belowdisplayskip}{2pt}
\mathcal{L}_{overall}  =  \mathcal{L}_{decom} + \mu \mathcal{L}_{LayerCon}(\textbf{\emph{B}},\textbf{\emph{R}}) + \sigma \mathcal{L}_{LocCon}(\textbf{\emph{B}},\textbf{\emph{O}}).
\end{equation}

\noindent
\textbf{Layer Contrastive}: First, the clean image \textbf{\emph{B}} and the rain layer \textbf{\emph{R}} are vastly different, in which the rain streaks are simple and directional line-pattern, while the natural images are complex yet meaningful structures such as edges and textures. The \emph{dissimilarity} between the \textbf{\emph{B}} and \textbf{\emph{R}}, as two different categories, can be well modelled by CL as negative pairs. And it is very reasonable to take the patches in the same image as the positive samples. The important sampling strategy and encoder in CL will be discussed in next subsection. Referring to the rain patches $\textbf{\emph{p}}_{\textbf{\emph{R}}_\textbf{\emph{i}}}$ as negatives, while the background patches as the positives $\textbf{\emph{p}}_{\textbf{\emph{B}}_\textbf{\emph{i}}}$, the contrastive loss between the two layers \textbf{\emph{B}} and \textbf{\emph{R}} can be formulated:
\begin{equation}
\begin{aligned}
\setlength{\abovedisplayskip}{2pt}
\setlength{\belowdisplayskip}{2pt}
\mathcal{L}_{LayerCon} = -\frac{1}{N_B}\sum_{k=1}^{N_B}\sum_{i=1}^{N_B}\frac{\textrm{exp}({\textbf{\emph{f}}}_{\textbf{\emph{B}}_\textbf{\emph{i}}}\cdot {\textbf{\emph{f}}}_{\textbf{\emph{B}}_\textbf{\emph{k}}}/\tau)}{\sum_{j=1}^{N_R}\textrm{exp}({\textbf{\emph{f}}}_{\textbf{\emph{B}}_\textbf{\emph{i}}}\cdot {\textbf{\emph{f}}}_{\textbf{\emph{R}}_\textbf{\emph{j}}}/\tau)}\\
-\frac{1}{N_R}\sum_{m=1}^{N_R}\sum_{j=1}^{N_R}\frac{\textrm{exp}({\textbf{\emph{f}}}_{\textbf{\emph{R}}_\textbf{\emph{j}}}\cdot {\textbf{\emph{f}}}_{\textbf{\emph{R}}_\textbf{\emph{m}}}/\tau)}{\sum_{i=1}^{N_B}\textrm{exp}({\textbf{\emph{f}}}_{\textbf{\emph{R}}_\textbf{\emph{j}}}\cdot {\textbf{\emph{f}}}_{\textbf{\emph{B}}_\textbf{\emph{i}}}/\tau)},
\end{aligned}
\end{equation}
where $\textbf{\emph{f}}_{\textbf{\emph{B}}_\textbf{\emph{i}}} = E_{D}(\textbf{\emph{p}}_{\textbf{\emph{B}}_\textbf{\emph{i}}})$, $\textbf{\emph{f}}_{\textbf{\emph{R}}_\textbf{\emph{j}}} = E_{D}(\textbf{\emph{p}}_{\textbf{\emph{R}}_\textbf{\emph{j}}})$, $\tau$ denotes the scale temperature parameter \cite{chen2020simple}. $E_{D}$ is the encoder of contrastive network. The features $\textbf{\emph{f}}_{\textbf{\emph{B}}_\textbf{\emph{k}}}$ are extracted from the non-local patches $\textbf{\emph{p}}_{\textbf{\emph{B}}_\textbf{\emph{k}}}$ of $\textbf{\emph{p}}_{\textbf{\emph{B}}_\textbf{\emph{i}}}$, while the $\textbf{\emph{f}}_{\textbf{\emph{R}}_\textbf{\emph{m}}}$ are extracted from the non-local patches $\textbf{\emph{p}}_{\textbf{\emph{R}}_\textbf{\emph{m}}}$ of $\textbf{\emph{p}}_{\textbf{\emph{R}}_\textbf{\emph{j}}}$. $N_B$ and $N_R$ denote the sample numbers of positives and negatives. The layer contrastive could facilitate us to better push the image layer away from rain layer, and pull each layer further to different clusters.

\noindent
\textbf{Location Contrastive}: Second, we can observe that the clean image \textbf{\emph{B}} and the observed image \textbf{\emph{O}} are visually close to each other, since the rain streaks \textbf{\emph{R}} are much simpler than \textbf{\emph{B}}. The \emph{similarity} between patches of the same location in \textbf{\emph{B}} and \textbf{\emph{O}}, as the same view, can be well modelled as the positive samples. Consequently, we set the patches with different locations as the negative samples. Note that, here for location contrastive, there should be only one positive sample, since the location correspondence is exactly one-to-one. The encoder of image generator $E_{G_{\textbf{\emph{B}}}}$ is utilized to extract the patch features, denoted as ${\textbf{\emph{v}}}_{\textbf{\emph{O}}_\textbf{\emph{i}}} = E_{G_{\textbf{\emph{B}}}}(\textbf{\emph{p}}_{\textbf{\emph{O}}_\textbf{\emph{i}}})$, and ${\textbf{\emph{v}}}_{\textbf{\emph{B}}_\textbf{\emph{i}}} = E_{G_{\textbf{\emph{B}}}}(\textbf{\emph{p}}_{\textbf{\emph{B}}_\textbf{\emph{i}}})$. Thus, the location contrastive loss is formulated as:
\begin{equation}
\begin{aligned}
\setlength{\abovedisplayskip}{2pt}
\setlength{\belowdisplayskip}{2pt}
&\resizebox{0.865\hsize}{!}{$\mathcal{L}_{LocCon} = \sum_{i=1}^{N}\frac{\textrm{exp}({\textbf{\emph{v}}}_{\textbf{\emph{O}}_\textbf{\emph{i}}}\cdot {\textbf{\emph{v}}}_{\textbf{\emph{B}}_\textbf{\emph{i}}}/\tau)}{\textrm{exp}({\textbf{\emph{v}}}_{\textbf{\emph{O}}_\textbf{\emph{i}}}\cdot {\textbf{\emph{v}}}_{\textbf{\emph{B}}_\textbf{\emph{i}}}/\tau)+   \sum_{j=1}^{N}\textrm{exp}({\textbf{\emph{v}}}_{\textbf{\emph{O}}_\textbf{\emph{j}}}\cdot {\textbf{\emph{v}}}_{\textbf{\emph{B}}_\textbf{\emph{i}}}/\tau)},$}
\end{aligned}
\end{equation}
where $N$ is the negative sample numbers. The location contrastive constrains the restored background patches $\textbf{\emph{p}}_{\textbf{\emph{B}}_\textbf{\emph{i}}}$ at location $i$ to be related (positive) with the corresponding input patches $\textbf{\emph{p}}_{\textbf{\emph{O}}_\textbf{\emph{i}}}$ in comparison to other random patches $\textbf{\emph{p}}_{\textbf{\emph{O}}_\textbf{\emph{j}}}$, so as to retain the image content.

\subsection{Non-local Sampling Strategy}
\label{sampling section}
In contrastive learning, the negatives are the samples which should be discriminated by the learned representations, while the positives are highly related and possess the invariance in the learned representations. The previous methods usually use the augmentations to construct the single instance positives and randomly sampling as the negatives \cite{chen2020simple}. Note that, the self-similarity is a generic and powerful prior knowledge. In this work, we introduce the non-local self-similarity to automatically select both positive and negative samples  within a single image. We employ the block matching \cite{dabov2007image} with $L_2$ Euclidian distance to measure the dissimilarity/similarity in image space:
\begin{equation}\label{eq:PatchMatching}
\setlength{\abovedisplayskip}{2pt}
\setlength{\belowdisplayskip}{2pt}
Dist(\textbf{\emph{p}}_{i}, \textbf{\emph{p}}_{{i}_{R}})  =  || \textbf{\emph{p}}_{i} - \textbf{\emph{p}}_{{i}_{\Omega}}||^2,
\end{equation}
where $\textbf{\emph{p}}_{i}$ is the query patch, $\textbf{\emph{p}}_{{i}_{\Omega}}$ are the searched patches in the support set $\Omega$. We take the top-\emph{k} smallest \emph{Dist}() as the similar patches, while the top-\emph{k} largest \emph{Dist}() can be regarded as the dissimilar patches. On one hand, the non-local sampling with similar structures would greatly ease the learning difficulty. On the other hand, the small perturbation within the similar samples would further improve the diversity. Moreover, the patches cropped from the image itself would provide more reliable representation learning. The non-local sampling strategy can be applied for sampling the positive and negative. Here we briefly describe how we use the non-local sampling in very flexible ways.

\begin{figure}[t]
  \centering
   \includegraphics[width=1.0\linewidth]{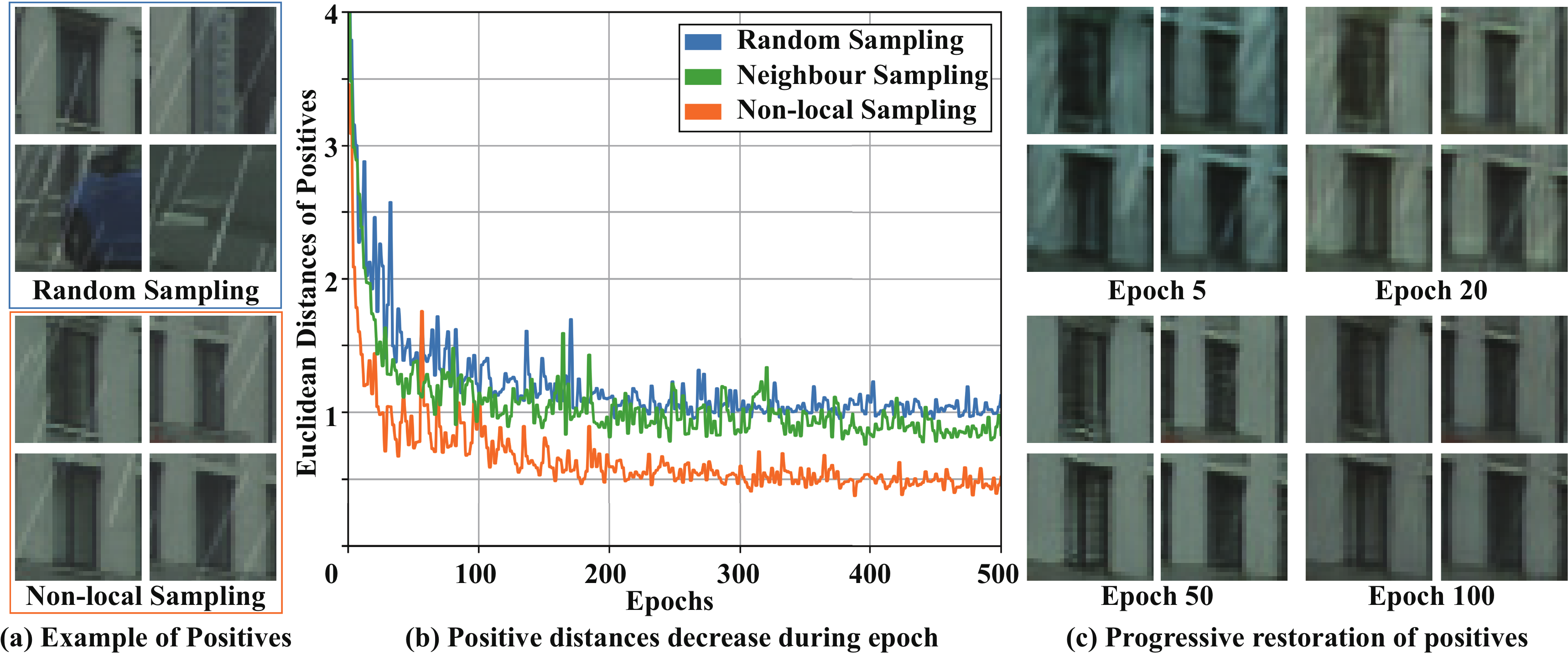}
  \caption{Effectiveness of the non-local sampling strategy. (a) Example of the randomly sampled positives with low similarity in comparison with the non-local self-similar sampled positives. (b) The Euclidean distances of positives decrease rapidly to a relative low level by non-local sampling strategy when compared with the random strategy, indicating the self-similarities are gradually learned and the patches are more relevant in the deraining procedure. (c) The similar patches guide each other to gradually restore the clean image and remove the randomly distributed rains.}
   \vspace{-10pt}
  \label{SamplingStrategy}
\end{figure}

\noindent
\textbf{Non-local Sampling in Layer Contrastive.} In layer contrastive, the clean image and rain streaks can be regarded as two distinct categories where they have intra-class similarity and inter-class dissimilarity. Our principle is that the positive samples (clean image patches in \textbf{\emph{B}}) should be pulled together as much as possible, so is the negative samples (rain streak patches in \textbf{\emph{R}}) which can also be pulled together. That is to say, we enforce the non-local sampling on both the positive and negative samples. Compared with single positive instance, the multiple non-local positive samples would benefit us to improve the feature representation. The recent research has also shown that positives from multiple instances could improve the representations if sampled appropriately (with supervised labels \cite{khosla2020supervised} or multiple modalities \cite{Han2020VideoRepresentation}). Moreover, compared with the random negative samples, the non-local sampling could additionally model the relationship within the samples.

To illustrate this, Figure \ref{SamplingStrategy} shows the superiority of the non-local sampling on positives. Figure \ref{SamplingStrategy}(a) shows an example of random and non-local sampled patches. The non-local patches possess the structure self-similarities in comparison with the random sampled ones. During training, we continually re-sample the non-local positives and calculate the similarity by Euclidean distance. Compared with random sampling or neighbour sampling which samples the surrounding patch neighbours, the distances of non-local positives decrease rapidly, and converge at a relatively low level, which indicates the self-similarities are gradually learned and the patches are more relevant in the restoration procedure [Fig. \ref{SamplingStrategy}(b)]. Therefore, by maximizing the correlations of positives, the self-similar patches guide each other to gradually restore the clean image and remove the rain streaks. The progressive deraining results are shown in Fig. \ref{SamplingStrategy}(c).

\noindent
\textbf{Non-local Sampling in Location Contrastive.} The observed image \textbf{\emph{O}} and clean image \textbf{\emph{B}} are very similar to each. In location contrastive, the goal is to retain the image content and remove the rain streaks in observed image, which is exactly a image-to-image translation task.
Thus, we follow the CUT \cite{park2020contrastive} by setting the patches of the same location in \textbf{\emph{B}} and \textbf{\emph{O}} as the positive samples with a large batch size. The previous methods including CUT randomly select the different patches as the negative. However, it is more reasonable that the more dissimilar from the positive sample, the better the negative sample is. This motivates us to still use the non-local sampling strategy to construct the negative patches. Instead of calculating the nearest top-\emph{k} samples, we choose the farthest top-\emph{k} samples (the largest distance) which means they are mostly different from the target positive. We name this negative sampling as the reverse non-local sampling.

\begin{figure}[t]
  \centering
  \includegraphics[width=1.0\linewidth]{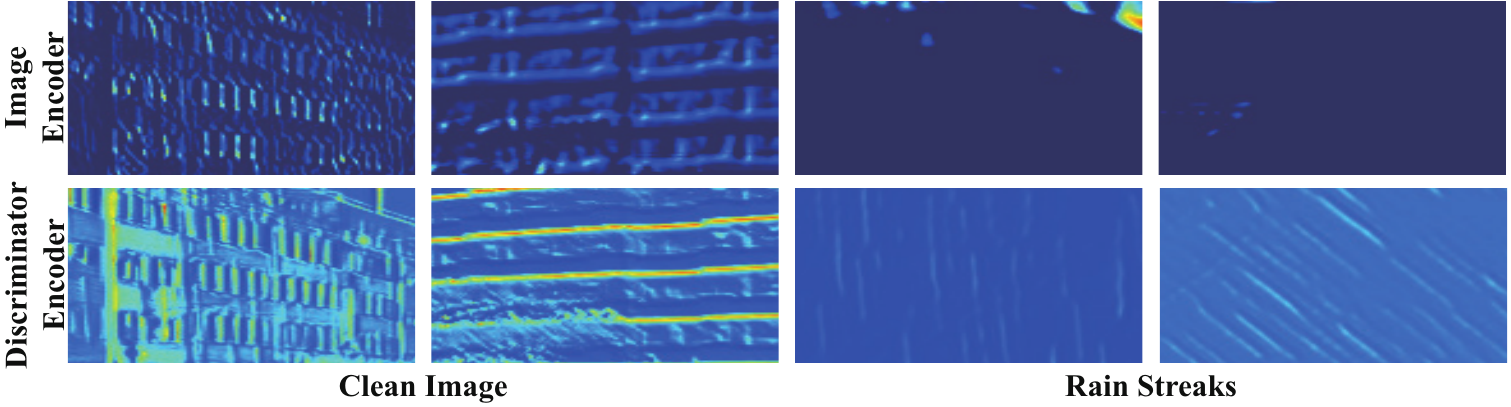}
  \caption{Effectiveness of the discriminator encoder. The first row shows the features extracted from image encoder. Although the extracted features in two different images are clear, the image generator has nearly no response to the rain streaks. The second row shows the features extracted from the discriminator encoder. The extracted features in two images and rain streaks are both clear and discriminative. This strongly supports the effectiveness of the discriminator serving as the encoder for the image and rain layer.}
   \vspace{-10pt}
  \label{FeatureEncoder}
\end{figure}

~\\[-30pt]

\subsection{Feature Encoder}
~\\[-30pt]

In contrastive learning, the feature encoder is to map the inputs to the embedding low-dimensional feature representation space that facilitates the measurement of the distances between positive and negative samples. It has been recognized that for different CL tasks, the choice of the encoder would vastly influence the final performance \cite{le2020contrastive}. In this work, we also demonstrate that the encoder is indeed tasks dependent for low-level restoration tasks, and explore different encoders for both the layer and location contrastive constraints intuitively and experimentally.

As for the layer contrastive, the goal is to differ the rain streaks from the clean image, which has been analyzed that this is analog to a classification problem. That is to say, the encoder of the layer contrastive should extract the high-level semantic about the category information. The discriminator is in line with the layer contrastive encoder, which can differ the image from non-image component including the rain streaks. As for the location contrastive, the clean image and the observed rainy image are very similar to each other, in which the clean image is the dominant component in rainy image. In other words, the encoder of the location contrastive should well extract the image features. The image generator can satisfactorily achieve this goal.

To verify our hypothesis, Figure \ref{FeatureEncoder} visualizes the embedded features map encoded by different encoders: image generator and discriminator. The first row shows the features extracted from image encoder, and the second row shows the features extracted from discriminator encoder. We select two different clean images and two different rain streaks as the example. We can observe that the image generator could effectively extract the image structures, while it cannot extract any informative information from the rain streaks. On the contrary, the line patterned rain streaks and the image structure can be clearly observed in the features extracted by the discriminator encoder. The discriminator focuses on the distinguishable features of image and non-image factors to perform the classification task, which matches the layer contrastive learning task better.

\begin{table}[]
  \centering
  \setlength{\abovecaptionskip}{1pt}
  \setlength{\belowcaptionskip}{1pt}
  \footnotesize
  \caption{Quantitative comparisons with SOTA unsupervised methods on synthetic and real datasets.}
  \renewcommand\arraystretch{1.05}
   \begin{tabular}{c|c|c|c|c|c|c}
   \hline
   \multirow{2}{*}{Methods} & \multicolumn{3}{c|}{RainCityscapes} & \multicolumn{3}{c}{SPA} \\
   \cline{2-7}
   & PSNR & SSIM & NIQE  & PSNR & SSIM & NIQE \\
   \hline
    DSC 			& 24.91 & 0.7603 & 6.17 & 33.71 & 0.9127 & 9.82 \\
   DIP 			& 22.45 & 0.6936	& 7.86 & 30.36 & 0.8422 & 9.97\\
   CycleGAN 	& 24.86 & 0.7906 &3.68 & 33.54 & 0.9127 & \textbf{6.67} \\
   UDGNet 	& 25.16 & \textbf{0.8749} &	5.31  & 29.67 & 0.9299 & 9.50 \\
   CUT 	& 25.21 & 0.8225 & 4.08 & 32.97 & 0.9434 & 9.60 \\
   NLCL 	& \textbf{26.46} & 0.8666 & \textbf{3.67} & \textbf{33.82} & \textbf{0.9468} & 9.55 \\\hline
  \end{tabular}
   \vspace{-10pt}
  \label{quanCompare}
 \end{table}

\section{Experiments}

\begin{figure*}[t]
\setlength{\abovecaptionskip}{2pt}
  \setlength{\belowcaptionskip}{-5pt}
  \centering
  {\includegraphics[width=1.0\linewidth]{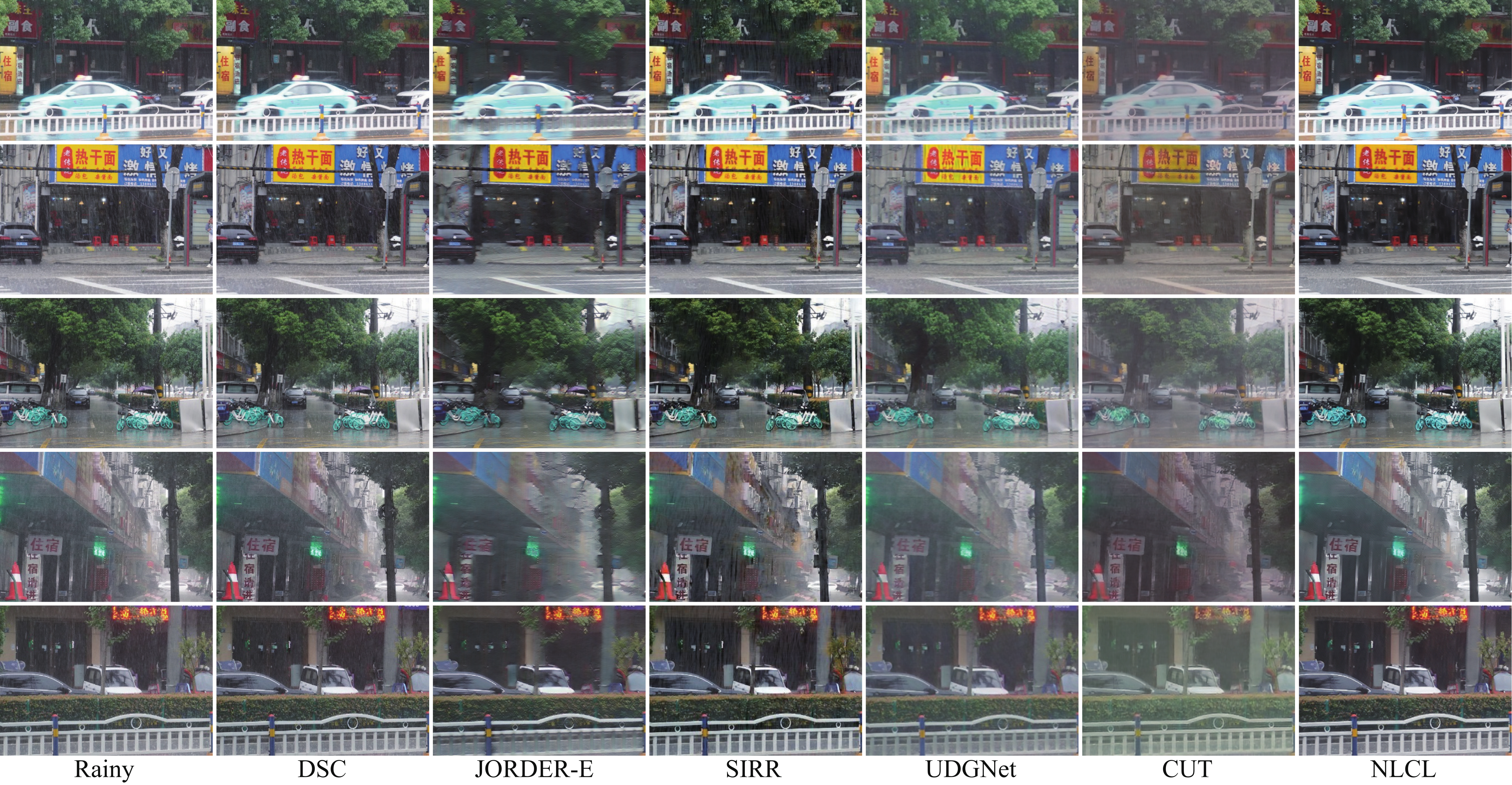}}
	\caption{Visual comparisons in real rainy scenes including both  rain streaks and heavy haze. We suggest to view the zoomed results on PC.}
	\label{Real}
\end{figure*}

\subsection{Implementation Details}
We utilize the same ResNet architectures \cite{johnson2016perceptual} for both the background extraction and rain extraction. Please refer to the supplementary for details. PatchGAN \cite{isola2017image} is employed for the discriminator. We first calculate the top-\emph{k} non-local patches in image space, then obtain the multilayer features \cite{park2020contrastive} from the encoder, and finally embed the non-local features through a two-layer MLP with 256 units. The sampling number $N, N_B$, and $N_R$ are set as 256, 8, 256. The encoder updating follows the setting of MoCo \cite{he2020momentum}, using momentum value $0.99$ and temperature $0.77$. The balance weights for each loss $\lambda$, $\delta$, $\mu$, $\sigma$ are set as $0.1, 1, 1, 1$. During the training, the original images are randomly cropped into $256 \times 256$ as input without any augmentation. We adopt the Adam optimizer and train the network with learning rate $0.0001$, and batch size $4$ on four RTX 2080TI GPUs.

\subsection{Datasets and Experimental Settings}

We conduct the experiments on both synthetic dataset RainCityscapes \cite{hu2019depth} and real dataset SPA \cite{wang2019spatial}. To simulated the real situation, we split the RainCityscapes with 1400 for training and 175 for testing. Note that we have no access to the ground truth and can only learn in an unsupervised manner. For the real dataset, we obtain 2000 rainy images from SPA for training and 200 rainy images for testing. For a fair comparison, we mainly select the unsupervised methods, including the optimization-based DSC \cite{luo2015removing}, CNN-based DIP \cite{ulyanov2018deep}, GAN-based CycleGAN \cite{zhu2017unpaired}, contrastive learning-based CUT \cite{park2020contrastive}, and optimization-driven deep CNN \cite{yu2021unsupervised}. Furthermore, we compare with state-of-the-art supervised JORDER-E \cite{yang2019joint} on the real rainy images. We employ the full-reference PSNR and SSIM to evaluate the deraining performance, and also the no-reference natural image quality evaluator (NIQE) \cite{mittal2012making} for comprehensive evaluation.

~\\[-25pt]

\subsection{Comparisons with State-of-the-arts}

~\\[-25pt]

In Table \ref{quanCompare}, we report the quantitative results on Cityscape and SPA. These datasets mainly contains the rain streaks with different visual appearances without the veiling in heavy rainy images. The quantitative results of NLCL mostly outperform state-of-the-art methods, which verifies the robustness of the proposed NLCL. Note that UDGNet mainly takes advantage of the directionality of the rain streaks, which is very suitable for the directional rain streaks in Cityscape. CycleGAN is an image generation method  aiming at visually natural appearance, which matches the goal  of NIQE, but cannot well preserve the original image content in terms of the relatively low PSNR. We do admit that our NLCL is not designed for the quantitative index on rain streaks. Instead, our philosophy is to unsupervisedly handle the real rains. To validate this, in Fig. \ref{Real}, we compare with the state-of-the-art on real-world rainy images, which contains both the rain streak and veiling. NLCL consistently achieves more natural and better visual results, which not only remove the rain streaks but also the veiling artifacts. The results strongly support the effectiveness of the CL and non-local sampling for better distinguishing real rain from image texture.

~\\[-25pt]

\subsection{Ablation Study}

~\\[-25pt]

\noindent
\textbf{The Effectiveness of the Non-local Sampling Strategy.} In Table \ref{ablationSamplingStrategy}, we compare the different sampling strategies for both the positives and negatives, including the random sampling, neighbour sampling (8 nearest neighbour patches), and the non-local sampling. These experiments are all performed on the layer contrastive. Compared with the random sampling, the non-local sampling for both the positive and negative could obviously improve the restoration results. That is to say, the non-local sampling is favorable to learn the image and rain streaks similarity, thus indeed reduces the variance within the positives and negatives, and at the same time enlarge the discrepancy between them. The neighbour sampling could slightly improve the results, while the non-local sampling still obtains the best performance.

 \begin{table}[t]
 \small
  \centering
   \setlength{\abovecaptionskip}{1pt}
  \setlength{\belowcaptionskip}{-2pt}
\caption{Ablation on different sampling strategies.}
  \begin{threeparttable}
  \renewcommand\arraystretch{1.0}
  \setlength{\tabcolsep}{1.6mm}{
  \begin{tabular}{c|cc|cc}
   \hline
   	 Positive & Negative & PSNR & SSIM  \\
	\hline	
	 Random	&	Random	& 25.83	&	0.8471	\\
	 Neighbour	&	Random	&	26.03	&	0.8491	\\
	 Neighbour	&	Neighbour	&	25.97	&	0.8477	\\
	 Non-local	&	Random	& 26.18	&	0.8489	\\
	 Random	& Non-local	&	26.16	&	0.8531	\\
	 Non-local	&	Non-local	&	\textbf{26.46}	&	\textbf{0.8666}	\\\hline
  \end{tabular}}
  \end{threeparttable}
   \vspace{-5pt}
  \label{ablationSamplingStrategy}
  \end{table}

\begin{table}[t]
 \small
\centering
 \setlength{\abovecaptionskip}{1pt}
  \setlength{\belowcaptionskip}{-2pt}
  \caption{The choice of different feature encoders.}
  \begin{threeparttable}
  \renewcommand\arraystretch{1.05}
  \setlength{\tabcolsep}{0.5mm}{
  \begin{tabular}{c|ccc}
  \hline
  Encoder & PSNR  & SSIM  & NIQE  \\
  \hline
  Image Generator & 24.86 & 0.8046 & 3.83\\
  Image-Rain Generator & 24.12 & 0.8023 & 3.95 \\
  Discriminator & \textbf{26.46}& \textbf{0.8666} & \textbf{3.67} \\\hline
  \end{tabular}}
  \end{threeparttable}
  \vspace{-5pt}
  \label{ablationFeatureEncoder}
 \end{table}

\noindent
\textbf{The Choice of Different Feature Encoders.} The choice of the encoder for latent feature space is very important. In Table \ref{ablationFeatureEncoder}, we test different encoders for layer contrastive feature embedding. First, we take the image generator as the feature encoder for both the image and rain layers. Second, we utilize the image generator and rain generator as the feature encoder for the image and rain layer, respectively. Third, we employ the discriminator as the feature encoder for both the image and rain layers. The discriminator encoder has achieved the best result, which verifies the discriminator is suitable to distinguish the image from rain patches.

  \begin{table}[t]
 \setlength{\abovecaptionskip}{1pt}
  \setlength{\belowcaptionskip}{-2pt}
 \small
  \centering
  \caption{The analysis of optimal sampling number.}
  \begin{threeparttable}
  \renewcommand\arraystretch{1.0}
  \setlength{\tabcolsep}{1.5mm}{
  \begin{tabular}{c|cccc}
  \hline
 \diagbox{Pos}{Neg} & 64 & 128 & 256 & 512 \\
 \hline
 4 		& 23.21 & 26.11 & 26.19 & 26.30\\
 8 		& 24.55 & 26.30 & \textbf{26.46} & 26.42 \\
 16 	& 24.54 & 25.96 & 26.02 & 26.11 \\
 32 	& 23.49 & 25.02 & 25.40 & 25.37\\\hline
  \end{tabular}}
  \end{threeparttable}
  \vspace{-5pt}
  \label{ablationSamplingSize}
 \end{table}

  \begin{table}[t]
\setlength{\abovecaptionskip}{1pt}
  \setlength{\belowcaptionskip}{0pt}
  \centering
   \small
  \caption{Different strategy ablations on location contrast.}
  \begin{threeparttable}
  \setlength{\tabcolsep}{1mm}{
  \begin{tabular}{c|ccc}
   \hline
   Ablations & PSNR & SSIM & NIQE\\
   \cline{1-4}
   Random Sampling & 26.33 & 0.8617 & 3.81\\
   \cline{1-4}
   Discriminator Encoder & 24.71 & 0.8476 & 3.94\\
   \cline{1-4}
   Sample Number 64 & 25.03 & 0.8646 &3.88 \\
   Sample Number 128 & 26.14 & 0.8604 & 3.75 \\
   Sample Number 512 & 25.98 & 0.8594 & 3.70\\
   \cline{1-4}
   NLCL & \textbf{26.46} & \textbf{0.8666} & \textbf{3.67}\\
    \hline
  \end{tabular}}
    \end{threeparttable}
    \vspace{-5pt}
  \label{LocationContrast}
 \end{table}

\noindent
\textbf{The Influence of the Non-local Sampling Number.} We show how the sampling numbers affect the derain result in Table \ref{ablationSamplingSize}. The PSNR increases when the positive sizes grow to an appropriate number, and then decrease since the excessive positives are somehow dissimilar. 8 positives and 256 negatives obtain the best performance. The reason is that most of the rain have the similar line patterns, thus more non-local similar patches can be found to boost the learning than complex image patches. Moreover, the sampling number is not the larger the better, since enforcing the dissimilar patches to be similar may violate the similar assumption.

\noindent
\textbf{The Strategies of Location Contrast.} We further study the strategies of location contrast in Table \ref{LocationContrast}, which shows the improvement from reverse non-local sampling. Moreover, the image generator encoder is much better than discriminator to preserve the image content in location contrastive. 256 is an appropriate number for the sampling number.

\noindent
\textbf{The Effectiveness of Each Loss.} In Table \ref{ablationLosses}, we show how each loss contributes to the final result. The $\mathcal{L}_{LocCon}$ and $\mathcal{L}_{LayerCon}$ aim to learn the correlations between the rainy-clean images, and the rain-image layers, which could greatly improve the deraining results. The self-consistency and adversarial loss are the baseline of our model. $\mathcal{L}_{1}$ sparse loss could slightly improve the performance.

  \begin{table}[t]
 \setlength{\abovecaptionskip}{1pt}
  \setlength{\belowcaptionskip}{-2pt}
 \small
 \centering
  \caption{Effectiveness of each loss in NLCL.}
  \begin{threeparttable}
  \renewcommand\arraystretch{1.0}
  \setlength{\tabcolsep}{1.4mm}{
  \begin{tabular}{c|ccc}
  \hline
  Model & PSNR & SSIM & NIQE \\
  \hline
  w/o $\mathcal{L}_{adv}$				& 21.55 & 0.7984	& 5.03 \\
  w/o $\mathcal{L}_{1}$ 		& 26.33 & 0.8566 & 3.74 \\
  w/o $\mathcal{L}_{LocCon}$ 		& 25.20 & 0.8469 & 3.85 \\
  w/o $\mathcal{L}_{LayerCon}$		& 24.12 & 0.8402	& 3.98 \\
%  w/o $\mathcal{L}_{Self}$	& 20.30 & 0.7012 & 5.25\\
  NLCL		 										& \textbf{26.46 }& \textbf{0.8666} & \textbf{3.67}\\ \hline
  \end{tabular}}
  \end{threeparttable}
   \vspace{-5pt}
  \label{ablationLosses}
 \end{table}

 \begin{table}[t]
\setlength{\abovecaptionskip}{1pt}
  \setlength{\belowcaptionskip}{3pt}
 \scriptsize
   \renewcommand\arraystretch{1}
 \caption{The model size and inference time under image $256*256$.}
 \setlength{\tabcolsep}{1.5mm}
\begin{tabular}{c|c|c|c|c|c|c}
\hline
Method   & DSC   & JORDER-E & CycleGAN & UDGNet & CUT & NLCL   \\ \hline
Size(MB) & --    & 16.7     & 45.6     & 5.7    &  45.6   & 2.6    \\ \hline
Time(s)  & 33.95 & 0.128    & 0.0144   & 0.0170 &   0.0135  & 0.0098 \\ \hline
\end{tabular}
 \vspace{-5pt}
\label{Time}
\end{table}

\begin{figure}[t]
\setlength{\abovecaptionskip}{3pt}
  \setlength{\belowcaptionskip}{0pt}
  \centering
  \includegraphics[width=1.0\linewidth]{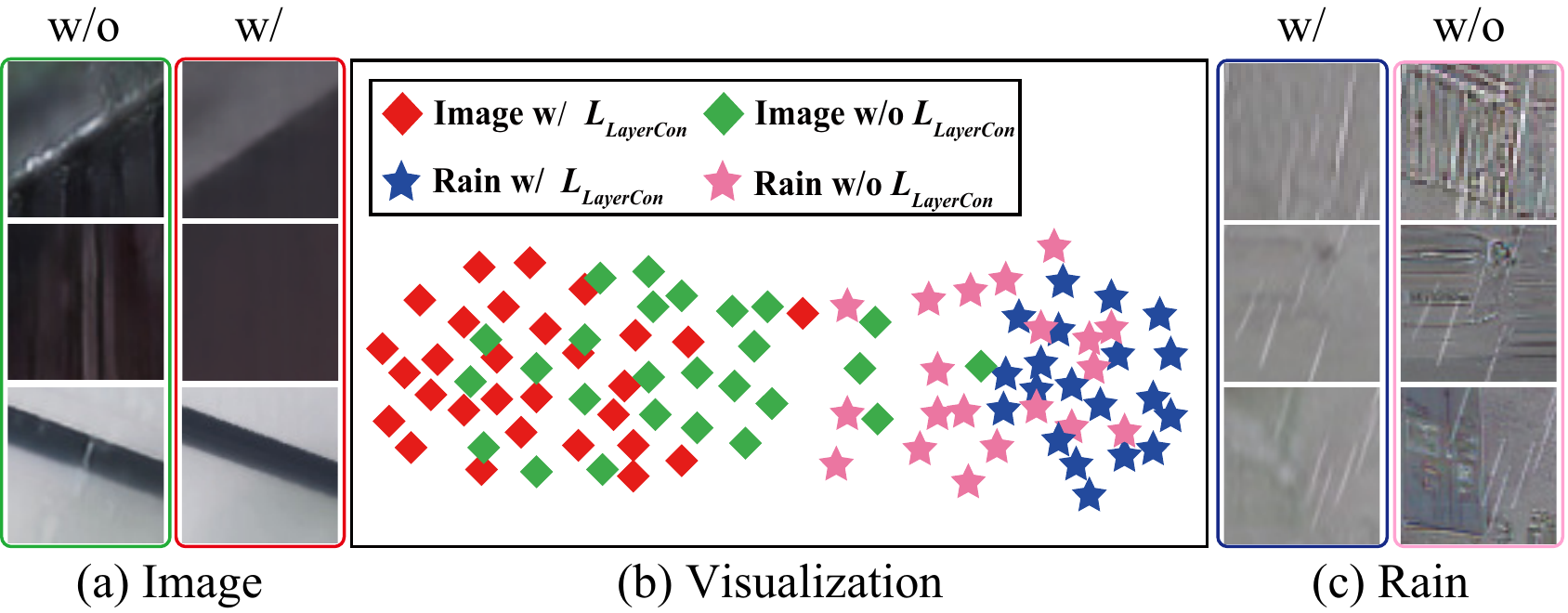}
  \caption{The effectiveness of the layer contrastive for better image and rain streaks decomposition. (a) and (c) show the decoupled rain and image patches w/ and w/o the layer contrastive. (b) visualizes the low-dimensional distributions w/ and w/o the layer contrastive.}
   \vspace{-10pt}
  \label{Decomposition}
\end{figure}

\subsection{Analysis and Discussion}
\noindent
\textbf{Effectiveness of Contrastive Learning.} In Fig. \ref{Decomposition}(b), we perform the tSNE to visualize the distribution of the decomposed image and rain layer w/ and w/o contrastive constraint. Without the layer contrastive, the distribution of the green rhombus (image) and the pink pentacle (rain streaks) are divergent. Moreover, they are mixed with each other which means they are still indistinguishable. On the contrary, with the layer contrastive, the distribution of the red rhombus (image) and the dark blur pentacle (rain streaks) are focused and distinguishable. In Fig. \ref{Decomposition}(a) and (c), with contrastive loss, the image and rain layers are better decoupled.

\begin{figure}
\setlength{\abovecaptionskip}{3pt}
  \setlength{\belowcaptionskip}{0pt}
  \centering
  {\includegraphics[width=1.0\linewidth]{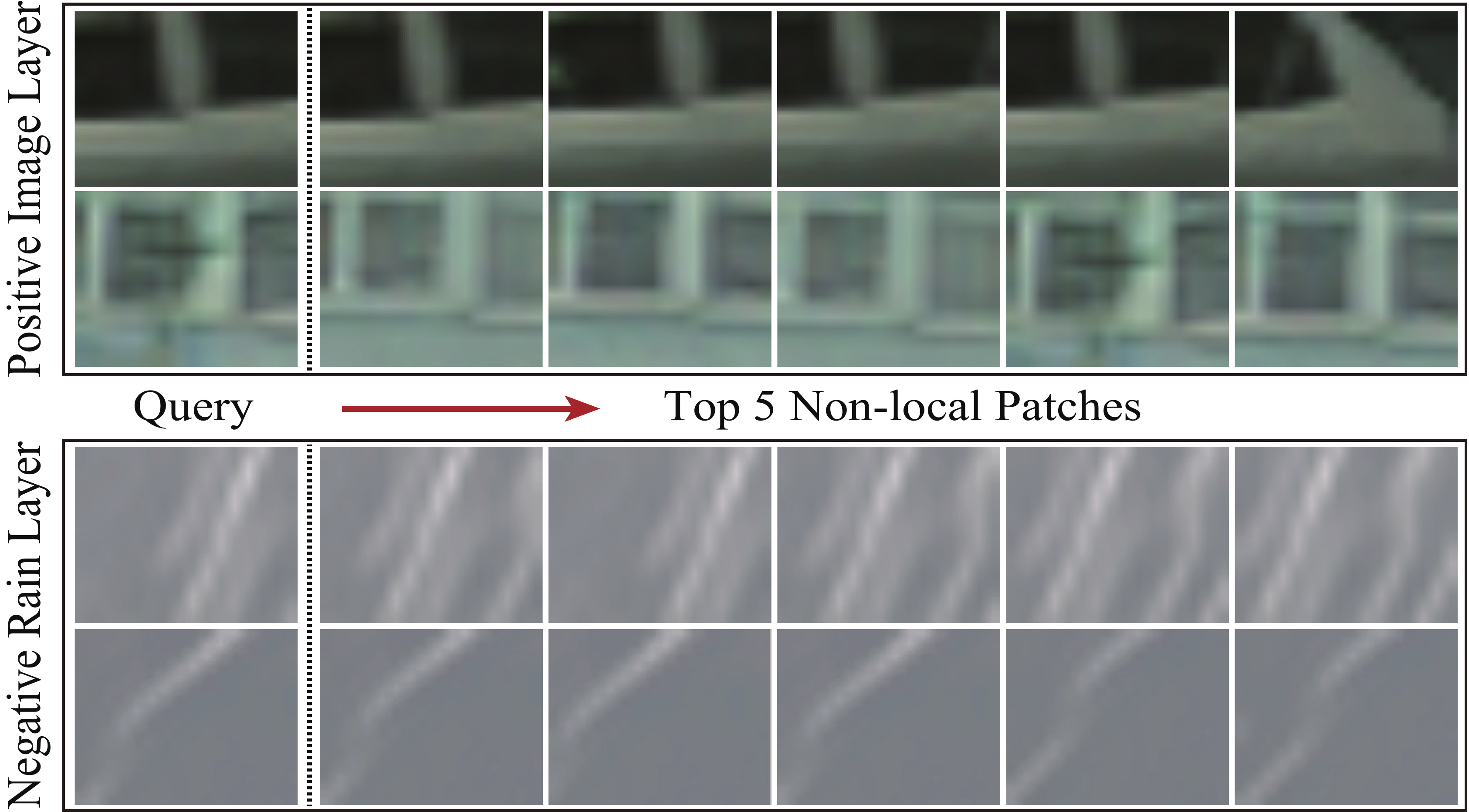}}
	\caption{The visualization of the Top5 non-local searched patches.}
	 \vspace{-5pt}
	\label{Non-local}
\end{figure}

\noindent
\textbf{Visualization of Self-similarity Patches.} We visualize the top 5 non-local positives and negatives of both light and heavy rain conditions in Fig. \ref{Non-local}. The extremely heavy rain would unavoidably increase the difficulty in non-local searching. The two-stage searching framework could be used, where the coarse clearer results are obtained before we search the non-local patches in the intermediated results. The positives and negatives are similar to that of the query key. The similarity is real and reliable with slight difference, instead of synthesis or fixed transformations. This intrinsic property facilitates us to learn discrimination representation.

\begin{figure}
\setlength{\abovecaptionskip}{3pt}
  \setlength{\belowcaptionskip}{0pt}
  \centering
  {\includegraphics[width=1.0\linewidth]{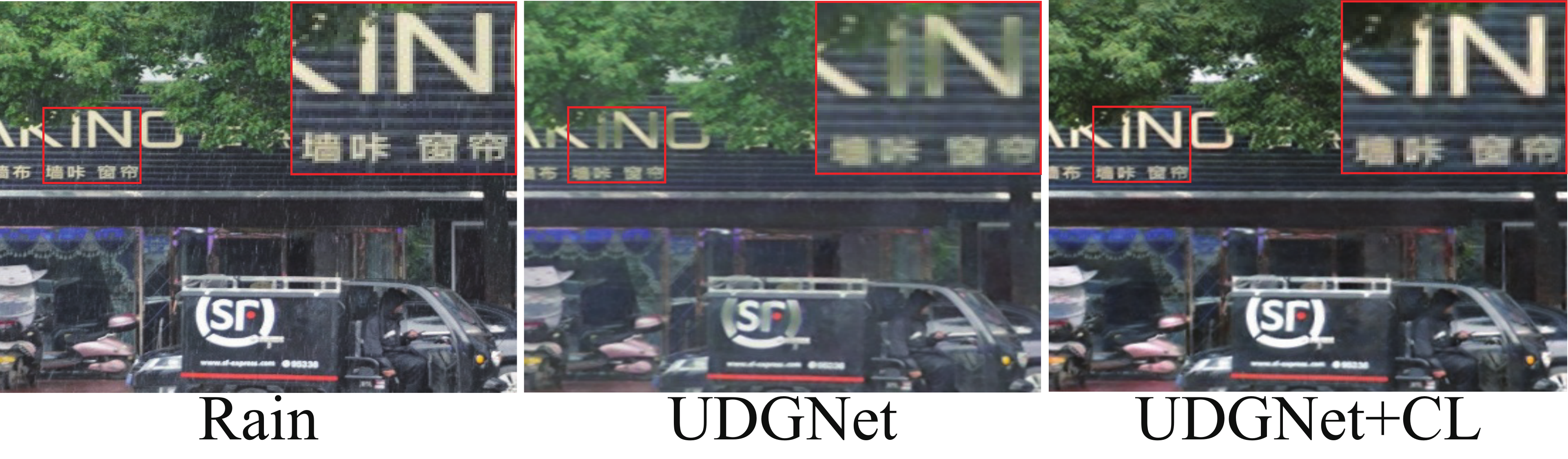}}
	\caption{The benefits of the NLCL when applied to UDGNet.}
	\label{ImprovedUDG}
	 \vspace{-10pt}
\end{figure}

\noindent
\textbf{The Benefit of the NLCL Strategy for Other Method.} Our NLCL is a general prior which can be naturally embedded into the existing methods for better decomposition. Here, we take the unsupervised deraining method UDGNet \cite{yu2021unsupervised} as example. As shown in Fig. \ref{ImprovedUDG}, although UDGNet could well remove the rain streaks without NLCL, the image structures have been unexpectedly removed along with rain. The result of UDGNet + NLCL is much better, such as the text.

\noindent
\textbf{Model Size and Running Time.} In inference phase, only the image generator is employed, making NLCL very fast and small, as shown in Table \ref{Time}. But in training phase, the additional nonlocal self-similarity searching is somewhat time-consuming. Normally, we take one day and a half for training 1400 images, which is the main limitation of NLCL. Speeding up the training time is one of our future work.

%It is known in MoCo \cite{he2020momentum} that the large the negatives number is (data diversity), the better the performance is. However, this is not true for non-local sampling, since enforcing the dissimilar patches to be similar may violate the similar assumption.

~\\[-32pt]

\section{Conclusion}

In this paper, we propose a novel non-local contrastive learning method, which explores the powerful self-similarity property within the image. Our method is totally unsupervised which can automatically decouple the image from the rain artifacts. We show that our non-local sampling strategy can be used to learn meaningful representations for both positives and negatives. Especially, the proposed non-local sampling strategy enriches the faithful, diverse and structural representation for both negatives and positives. Moreover, we provide an guidance of how to select an appropriate encoder for better feature embedding. Extensive experiments demonstrate the effectiveness of the proposed method.

\textbf{Acknowledgements.} This work was supported by National Natural Science Foundation of China under Grant No. 61971460 and No. 62101294, and Equipment Pre-Research Foundation under Grant No. 6142113200304.

%%%%%%%%% REFERENCES
{\footnotesize
\bibliographystyle{ieee_fullname}
\bibliography{egbib}
}

\end{document}